\pdfoutput=1
\documentclass{article}
\usepackage[preprint]{spconf}
\usepackage{amsmath}
\usepackage{amssymb}
\usepackage{amsfonts}
\usepackage{bbding}
\usepackage{booktabs}
\usepackage{enumitem}
\usepackage{multirow}
\usepackage{tikz}
\usepackage{url}
\usepackage{color}
\usepackage{adjustbox}
\usepackage{tabulary}
\usepackage{graphicx}
\usepackage{algorithm}
\usepackage{algorithmicx}
\usepackage[normalem]{ulem}
\useunder{\uline}{\ul}{}
\usepackage[noend]{algpseudocode}

\copyrightnotice{\copyright\ IEEE 2019}
\toappear{To appear in {\it Proc.\ ICASSP 2019, May 12-17, 2019, Brighton, U.K.}}

\algnewcommand{\parState}[1]{\State%
	\parbox[t]{\dimexpr\linewidth-\algmargin}{\strut #1\strut}}

\usepackage[disable]{todonotes} %

\newcommand{\etal}{\textit{et al}.\@ }
\newcommand{\ie}{\textit{i}.\textit{e}.\@ }
\newcommand{\eg}{\textit{e}.\textit{g}.\@ }
\newcommand{\etc}{\textit{etc}.\@}

\newcommand{\norm}[1]{\left\lVert#1\right\rVert}

\title{Investigating the effects of word substitution errors on sentence embeddings\thanks{Personal use of this material is permitted. However, permission to reprint/republish this material for advertising or promotional purposes or for creating new collective works for resale or redistribution to servers or lists, or to reuse any copyrighted component of this work in other works, must be obtained from the IEEE. Contact: Manager, Copyrights and Permissions / IEEE Service Center / 445 Hoes Lane / P.O. Box 1331 / Piscataway, NJ 08855-1331, USA. Telephone: + Intl. 908-562-3966}}
\name{Rohit Voleti$^{1}$, Julie M. Liss$^{2}$, Visar Berisha$^{1,2}$
}
\address{Arizona State University
	  \\ Department of Electrical, Computer, \& Energy Engineering$^{1}$
	  \\ Department of Speech \& Hearing Science$^{2}$
	  \\ Tempe, AZ, USA
}

\begin{document}
\ninept

\maketitle

\begin{abstract}
	A key initial step in several natural language processing (NLP) tasks involves embedding phrases of text to vectors of real numbers that preserve semantic meaning. 
	To that end, several methods have been recently proposed with impressive results on semantic similarity tasks. 
	However, all of these approaches assume that perfect transcripts are available when generating the embeddings. 
	While this is a reasonable assumption for analysis of written text, it is limiting for analysis of transcribed text. In this paper we investigate the effects of word substitution errors, such as those coming from automatic speech recognition errors (ASR), on several state-of-the-art sentence embedding methods. 
	To do this, we propose a new simulator that allows the experimenter to induce ASR-plausible word substitution errors in a corpus at a desired word error rate. 
	We use this simulator to evaluate the robustness of several sentence embedding methods. 
	Our results show that pre-trained neural sentence encoders are both robust to ASR errors and perform well on textual similarity tasks after errors are introduced.
	Meanwhile, unweighted averages of word vectors perform well with perfect transcriptions, but their performance degrades rapidly on textual similarity tasks for text with word substitution errors. 
\end{abstract}


\section{Introduction \& Related Work}\label{sec:intro}
Many real-world applications motivate the need to accurately capture the semantic content of a sentence.
Examples include sentiment analysis of product reviews, customer service chatbots, biomedical informatics, among several others.
\emph{Word embeddings} map words from a lexicon to a continuous vector space in which nearby vectors are also semantically related.
Similarly, \emph{sentence embeddings} map individual phrases or sentences to a continuous vector space that preserve the text semantics.
The approaches to the word-embedding problem range from simple singular value decomposition of co-occurrence matrices~\cite{landauerSolutionPlatoProblem1997} to neural network models trained on large corpora (\eg \emph{word2vec}~\cite{mikolovEfficientEstimationWord2013}, \emph{GloVe}~\cite{penningtonGloveGlobalVectors2014}, and \emph{FastText}~\cite{bojanowskiEnrichingWordVectors2016}).

These approaches have revolutionized NLP research by showing impressive results on downstream NLP tasks; however, to the best of our knowledge, all of the previous work on sentence and word embeddings is built upon the assumption that the available text for training and testing each embedding model is perfectly transcribed.
In most real-world applications, it is unlikely that textual language data will be free of error.
In fact, an increasing number of applications rely on \emph{automatic speech recognition} (ASR) systems for transcriptions.
The performance of an ASR system can be characterized by its \emph{word-error rate} (WER), which defines the percentage of incorrect word errors given by the output of a particular system.
Typical modern ASR systems have a WER ranging from $\sim$10\% to $\sim$ 35\% \cite{bohoutaComparingSpeechRecognition2017}.
With a few exceptions, \ie \cite{liSpokenSQuADStudy2018a}, \cite{simonnetSimulatingASRErrors2018}, \cite{stuttleFrameworkDialogueData2004}, \cite{jungIntegratedDialogSimulation2008}, the effects of ASR errors have been largely ignored in many NLP applications.
And, to the best of our knowledge, no previous work has been conducted to evaluate the effects of ASR errors on sentence embeddings and their performance in downstream NLP tasks.

In this work, we evaluate the robustness of several state-of-the-art sentence embeddings to word substitution errors typical of ASR systems\footnote{WER calculation includes unintended word \textit{insertions}, \textit{deletions}, and \textit{substitutions}. We note that a limitation of our model is that it only considers potential substitution errors when simulating ASR error}.
To do this, we propose a new method for simulating realistic ASR transcription errors with a specified WER that is implemented with only publicly available tools for acoustic and semantic modeling.
We evaluate the resultant embeddings on the semantic textual similarity (STS) task, a popular research topic in NLP within the area of statistical distributional semantics.
In STS, the goal is to develop sentence embeddings that can successfully model the semantic similarity between two sentences (or another arbitrary collection of words). 
Several recently developed sentence embedding methods have shown very promising results on STS tasks \cite{mikolovEfficientEstimationWord2013}, \cite{penningtonGloveGlobalVectors2014}, \cite{kusnerWordEmbeddingsDocument}, \cite{aroraSimpleToughtoBeatBaseline2017}, \cite{muRepresentingSentencesLowRank2017}, \cite{conneauSupervisedLearningUniversal2017}, \cite{pagliardiniUnsupervisedLearningSentence2017}, \cite{ethayarajhUnsupervisedRandomWalk2018}; however, all have been evaluated using perfect transcripts.
We attempt to re-evaluate the results on standard STS datasets after introducing the errors simulated using our approach.
In short, the contributions of this work are: $1$) a new simulator for introducing ASR-plausible word substitution errors that utilizes phonetic and semantic information to randomly replace words in a corpus with likely confusion words, $2$) an evaluation of five recent sentence embedding methods and their robustness to simulated ASR noise, and $3$) an evaluation of the STS performance of these sentence embeddings with simulated ASR errors and a variable WER using the \emph{SICK}~\cite{marelliSemEval2014TaskEvaluation2014} and \emph{STS-benchmark}~\cite{cerSemEval2017TaskSemantic2017} datasets.

\section{Word Substitution Error Simulation}\label{sec:asr}
\begin{table}[t]
	\centering
	\resizebox{0.8\columnwidth}{!}{%
		\begin{tabular}{@{}|l|l|@{}}
			\toprule
			\multicolumn{1}{|c|}{\textbf{Original Sentence}}                                                                                          & \multicolumn{1}{c|}{\textbf{Corrupted Sentence}}                                                                                            \\ \midrule
			\begin{tabular}[c]{@{}l@{}}Obama holds out over \\ Syria strike.\end{tabular}                                                             & \begin{tabular}[c]{@{}l@{}}Obama \emph{helps} out \emph{every} \\ \emph{Sharia} strike.\end{tabular}                                                             \\ \midrule
			\begin{tabular}[c]{@{}l@{}}Russia warns Ukraine \\ against EU deal.\end{tabular}                                                          & \begin{tabular}[c]{@{}l@{}}Russia warns \emph{Euro} \\ against EU deal.\end{tabular}                                                               \\ \midrule
			\begin{tabular}[c]{@{}l@{}}Gov. Linda Lingle and \\ members of her staff \\ were at the Navy base \\ and watched the launch.\end{tabular} & \begin{tabular}[c]{@{}l@{}}Gov. \emph{Cindy} Lingle \emph{add}\\ \emph{mentors} of her \emph{staffs} were \\ at the \emph{NASA} base \\ and watched the \emph{launcher}.\end{tabular} \\ \midrule
			\begin{tabular}[c]{@{}l@{}}I have had the same \\ problem.\end{tabular}                                                                   & \begin{tabular}[c]{@{}l@{}}\emph{Eyes} have had the same \\ \emph{progress}.\end{tabular}                                                                 \\ \midrule
			\begin{tabular}[c]{@{}l@{}}A white cat looking \\ out of a window.\end{tabular}                                                           & \begin{tabular}[c]{@{}l@{}}A white cat \emph{letting} \\ out of a window.\end{tabular}                                                             \\ \bottomrule
		\end{tabular}%
	}
	\caption{Example sentence pairs from STS-benchmark~\cite{cerSemEval2017TaskSemantic2017} and SICK corpora \cite{marelliSICKCureEvaluation} after corrupting all sentences with WER of $30\%$. Substituted word errors are shown in italics. A high WER is used here to demonstrate the types of substitution errors simulated by our method, incorporating both semantic and phonemic distance measures.}
	\label{table:asr_errors}
\end{table}
In this section we propose a new word substitution error simulator intended to model plausible substitutions that an ASR algorithm might produce. 
Our approach is based on the observation that the nature of word substitution errors in ASR systems depends on the phonemic distance between the true word and the substituted word (because of the underlying acoustic model) \emph{and} on the semantic distance between the true word and the substituted word (because of the underlying language model). 
To that end, we define the probability of substituting word $w_i$ with word $w_j$ by 
\begin{equation}\label{eq:prob_cond}
	P_{\mathrm{subs}}( w_j | w_i) = \alpha \cdot \mathrm{exp}({-\frac{d_{ij}}{\sigma^2}}), 
\end{equation}
where $d_{ij}$ is a notion of distance between $w_i$ and $w_j$ comprised of both the phonemic and semantic distance, $\sigma$ is a user-defined parameter that controls the shape of the resulting probability mass function (PMF), and $\alpha$ is a normalization constant that makes the marginal PMF in Equation~\ref{eq:prob_cond} sum to one for each given $w_i$.
\\ \\
\noindent{\bf Estimating the substitution probabilities:} Given a corpus for which we want to simulate word substitution errors, we first compute the set of all unique words.
Next, we consider the pair-wise substitution error probabilities using Eqn. (\ref{eq:prob_cond}). Estimating the probability of a substitution requires that we estimate $d_{ij}$. 
Loosely speaking, we model the total distance as being comprised of a phonemic distance between the words (contribution of acoustic model in ASR) and a semantic distance between words (contribution of the language model in ASR). 

To estimate the phonemic distance, we use a phonological edit distance between words $w_i$ and $w_j$, $d_{ij}^{P}$ \cite{sandersPhonologicalDistanceMeasures2009}, \cite{allenLearningAlternationsSurface2015}, \cite{hallPhonologicalCorpusTools2015}, loosely based on the Levenshtein edit distance \cite{levenshteinBinaryCodesCapable1966}, which compares the number of single-character edits one string would need to be identical to another string.
We consider ARPABET transcriptions based on the \emph{CMU Pronouncing Dictionary} \cite{weideCMUPronouncingDictionary1998} to similarly compute phonemic similarity.
To encode each phoneme, we use the \emph{articulation features} provided by Hayes in~\cite{hayesIntroductoryPhonology2009}. 
The result is a binary feature matrix for each English phoneme in ARPABET. 
The phonological edit distance between two words can be computed as the number of \emph{single-feature} edits that are required to pronounce the first word like the second, as outlined by Sanders \etal in \cite{sandersPhonologicalDistanceMeasures2009}.

To estimate the semantic distance between the words, we use the \emph{GloVe} embeddings~\cite{penningtonGloveGlobalVectors2014} for every word in the corpus and estimate the pairwise \emph{cosine distance} as 
\begin{equation}\label{eq:cos_dist}
	d_{ij}^{S} = 1 - \cos{\theta_{ij}} = 1 - \frac{\mathbf{w_1}^T\mathbf{w_2}}{\left\lVert{\mathbf{w_1}}\right\rVert_2\left\lVert\mathbf{w_2}\right\rVert_2}
\end{equation}
where $\mathbf{w_i}$ and $\mathbf{w_j}$ represent the vector representations of two distinct words $w_i$ and $w_j$, and $\theta_{ij}$ represents the angle between the vectors.
\\ \\
\noindent{\bf Algorithm implementation:} The total distance in Equation~\ref{eq:prob_cond} can be modeled using some function of the two contributions discussed above, $d_{ij} = f(d_{ij}^{S}, d_{ij}^{P})$.
However, this approach requires that we estimate the conditional probability in Equation~\ref{eq:prob_cond} for every pair of words in a corpus; for large, realistic vocabulary sizes, this becomes prohibitively large.

To alleviate the need to estimate all pairwise probabilities, we only consider the $N=1000$ semantically most similar words in the corpus using $d_{ij}^{S}$ and estimate the marginal distribution for that subset of words, assuming that it is zero for all others. 
In addition, in Equation~\ref{eq:prob_cond}, we model $d_{ij}$ using only the contribution from the phonological edit distance. The parameter $\sigma$ can be chosen and tuned based on empirical results.
We found that setting $\sigma$ equal to the average phonological edit distance between each cluster of potential replacement words and the target word provided reasonable results. 
The overall procedure is summarized in Algorithm~\ref{alg:asr_error}.
\begin{algorithm}
	\caption{Random replacement of words in a given a corpus with a specified WER to simulate realistic ASR errors.}
	\label{alg:asr_error}
	\begin{algorithmic}[1]
		\Procedure{Corrupt Sentences}{$\mathrm{corpus}$, $\mathrm{WER}$}
			\State Find all unique tokens, $w_i$, in the $\mathrm{corpus}$ that exist in the set \hspace*{0.5cm} of pre-trained \emph{GloVe} embeddings
			\State Filter all $w_i$ to those in pronouncing dictionary
			\For{each $w_i$}
				\State Find $w_j$, $j = 1, \cdots, N$ most similar words by $d_{ij}^{S}$
				\State ARPABET transcription for $w_i$, all $w_j$ \Comment{CMU Dict}
				\For{each $w_j$}
					\State Compute $d_{ij}^P$ from $w_i$ to $w_j$, where $j = 1, \cdots, N$
				\EndFor
				\State Keep only $M$ values of $d_{ij}^P \leq \mathrm{thresh}$, where $M \leq N$
				\For{$j = 1, \cdots, M$}
					\State Compute $P_{\text{subs}}(w_j|w_i)$ \Comment{Eq.~\ref{eq:prob_cond}}\label{line:prob}
				\EndFor				
			\EndFor
			\State Randomly select words to replace given $\mathrm{WER}$
			\State Replace selected words with error words based on the \hspace*{0.5cm}probability distributions computed \Comment{Line \ref{line:prob}}
		\EndProcedure
	\end{algorithmic}
\end{algorithm}

In Table~\ref{table:asr_errors}, we provide several examples of the substitution errors simulated at a given WER of $30\%$.

\section{Sentence Embedding Methods}\label{sec:embeddings}

The sentence embedding methods described in this section have all been shown to perform well on STS tasks~\cite{piersmenComparingSentenceSimilarity2018}, \cite{peroneEvaluationSentenceEmbeddings2018} and serve as a representative set of models to evaluate robustness to ASR errors. 
A brief description of each method is provided below:
\\ \\
\noindent \emph{Simple Unweighted Average:} A common sentence embedding implementation is a computation of the arithmetic mean for all word vectors that comprise a sentence. This serves as a simple but effective baseline with pre-trained \emph{word2vec} embeddings~\cite{mikolovEfficientEstimationWord2013}. Additionally, averages can be computed after removing stop words which contain little semantic content (\eg "is", "the", \etc).
\\ \\
\noindent \emph{Smooth Inverse Frequency (SIF):} Arora \etal propose SIF embeddings~\cite{aroraSimpleToughtoBeatBaseline2017}, which involve two major components. First, a weighted average of the form $\frac{a}{a+p(w)}$ is computed, in which $a$ is a scalar value (a hyperparameter, tuned to $0.001$) and $p(w)$ is the probability that a word appears in a given corpus. This weighting scheme de-emphasizes commonly used words (with high probability) and emphasizes low probability words that likely carry more semantic content. Additionally, SIF embeddings attempt to diminish the influence of semantically meaningless directions common to the whole corpus. To do so, all word vectors in a dataset are concatenated into a matrix from which the first principal component is removed from each weighted average.
\\ \\
\noindent \emph{Unsupervised Smooth Inverse Frequency (uSIF):} Ethayarajh proposes a refinement to SIF known as uSIF, which claims improvements in many tasks (including STS)~\cite{ethayarajhUnsupervisedRandomWalk2018}. uSIF differs from SIF in that the hyperparameter $a$ is directly computed (and not tuned), making it fully unsupervised. Additionally, the first $m$ ($m=5$) principal components, each weighted by the factor $\lambda_1, \cdots, \lambda_m$ are subtracted for the common component removal step. 
Here, $\lambda_i = \frac{\sigma_i^2}{\sum_{i=1}^{m}{\sigma_i^2}}$, where $\sigma_i$ is the $i$-th singular value of the embedding matrix.
\\ \\
\noindent \emph{Low-Rank Subspace}: Mu \etal propose a unique sentence embedding in which sentences are represented by an $N$-dimensional subspace rather than a single vector \cite{muRepresentingSentencesLowRank2017}. Given word vectors of dimension $d$ and subspace rank of $N$, a sentence matrix is first constructed by concatenating word vectors and has dimension $d \times N$ (we use $d=300$ and $N=4$). Then, principal component analysis (PCA) is performed to identify the first $N$ principal components whose span comprise a rank-$N$ subspace in ${\rm I\!R}^d$. We consider this method for our simulated ASR error analysis to test whether the subspace representation is more robust to ASR errors than a vector representation.
\\ \\
\noindent \emph{InferSent:} Conneau \etal developed the \emph{InferSent} encoder that utilizes a transfer learning approach \cite{conneauSupervisedLearningUniversal2017}. The encoder is trained with a bidirectional LSTM neural network on the Stanford Natural Language Inference (SNLI) dataset, a labeled dataset that is designed for textual entailment tasks. The embeddings learned from the NLI task are then used to perform textual similarity tasks in STS.
\\ \\
\noindent \textbf{Computing Similarities:} Sentences represented by vectors (\ie averages, SIF, uSIF, \emph{InferSent}) can be compared with \emph{cosine similarity}, closely related to $d_{ij}^{S}$ in Equation~\ref{eq:cos_dist}.
Cosine similarity is given as $ \mathrm{CosSim} = 1 - d_{ij}^{S} = \cos{\theta_{ij}} = \frac{\mathbf{w_1}^T\mathbf{w_2}}{\norm{\mathbf{w_1}}_2\norm{\mathbf{w_2}}_2}$.
For subspace similarity, the authors in~\cite{muRepresentingSentencesLowRank2017} suggest the analogous concept of computing the \emph{principal angle} between the rank-$N$ subspaces for two sentences.
This can be readily obtained from the singular value decomposition. 
If we let the matrices $U(s_1)$ and $U(s_2)$ have columns that each contain the first $N$ principal components for sentences $s_1$ and $s_2$, the principal angle similarity given by:
\begin{equation}\label{eq:princ_angle}
	\mathrm{PrincAng}(s_1, s_2) = \sqrt{\sum\nolimits_{t=1}^{N}\sigma_t^2}
\end{equation}
In Equation~\ref{eq:princ_angle}, $\sigma_t$ represents the $t$-th singular value of the product $U(s_1)^TU(s_2)$.

\section{Results \& Discussion}\label{sec:results}

\begin{figure}[]
	\centering
	\includegraphics[width=0.86\columnwidth]{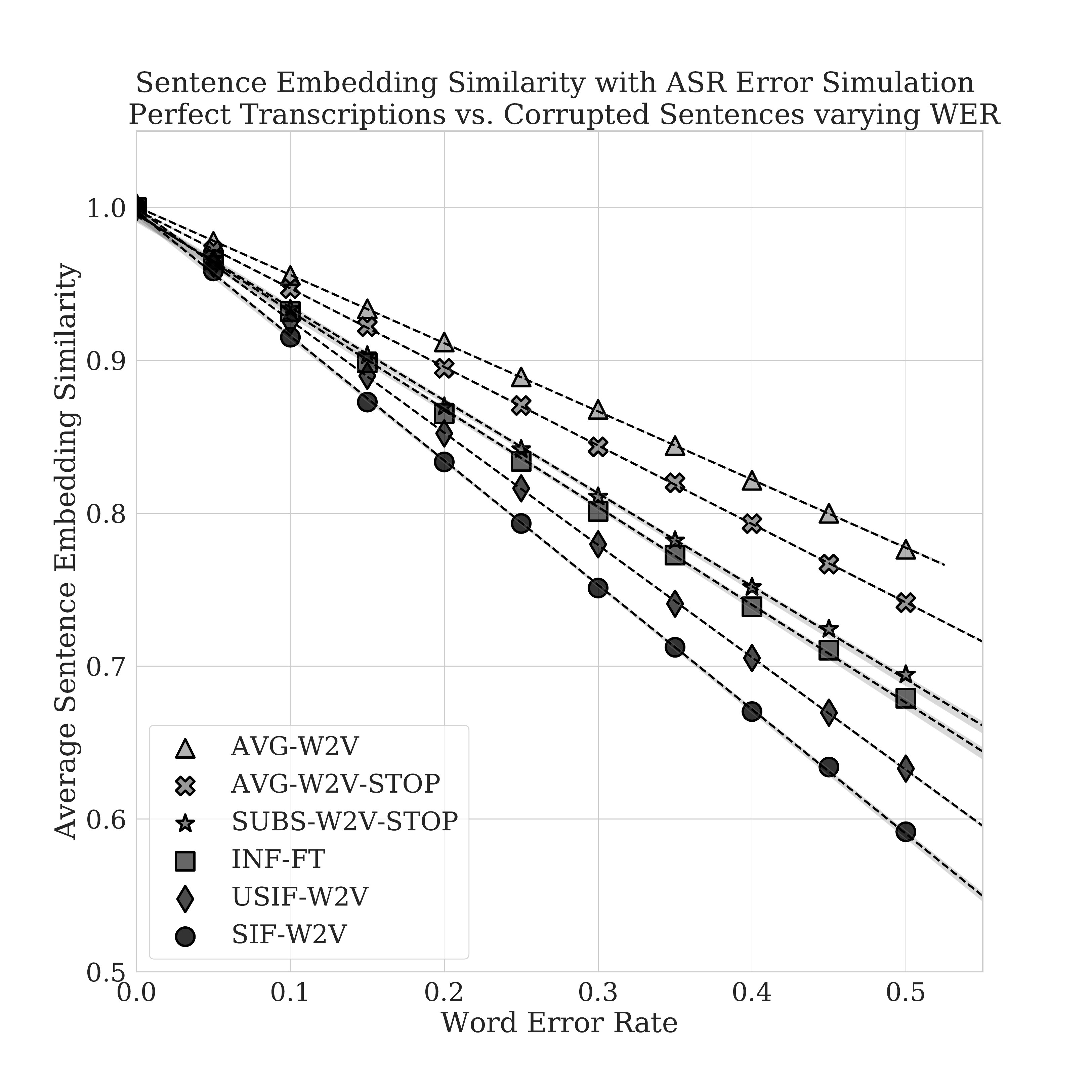}
	\caption{Regression plots for sentence embedding methods described in Section~\ref{sec:embeddings} as the WER is varied from $0\%$ to $50\%$. We consider averaging \emph{word2vec} vectors (\text{$\bigtriangleup$}), averaging \emph{word2vec} and removing stop words (\text{$\mathbb{X}$}), low-rank subspace representations with \emph{word2vec} and stop-words removed (\text{\FiveStarOpen}) \cite{muRepresentingSentencesLowRank2017}, \emph{InferSent} with \emph{FastText} embeddings ($\Box$) \cite{{conneauSupervisedLearningUniversal2017}}, SIF with \emph{word2vec} \cite{aroraSimpleToughtoBeatBaseline2017} (\text{$\bigcirc$}), and uSIF with \emph{word2vec} ($\diamondsuit$) \cite{ethayarajhUnsupervisedRandomWalk2018}.}
	\label{fig:asr_sentence_embeddings}
\end{figure}

\begin{table}[t]
	\centering
	\resizebox{\columnwidth}{!}{%
		\begin{tabular}{@{}|l|r|c|c|@{}}
			\toprule
			\multicolumn{1}{|c|}{\textbf{Sentence Embedding}} & \multicolumn{1}{c|}{\textbf{\begin{tabular}[c]{@{}c@{}}STS Corpus \\ (dev \& test set)\end{tabular}}} & \begin{tabular}[c]{@{}c@{}}$\mathbf{PCC}_{\mathbf{0\%}}$ / $\mathbf{PCC}_{\mathbf{10\%}}$ / $\mathbf{PCC}_{\mathbf{30\%}}$\\ ($\times 100$)\end{tabular} & $\frac{\mathbf{PCC}_{\mathbf{30\%}}}{{\mathbf{PCC}_{\mathbf{0\%}}}}$ \\ \midrule
			AVG-W2V:                                          & \begin{tabular}[c]{@{}r@{}}SICK:\\ STS-benchmark:\end{tabular}                                        & \begin{tabular}[c]{@{}c@{}}72.84 / 64.44 / 49.18\\ 67.40 / 59.23 / 45.64\end{tabular}                                                                    & \begin{tabular}[c]{@{}c@{}}67.52\%\\ 67.72\%\end{tabular}            \\ \midrule
			AVG-W2V-STOP:                                     & \begin{tabular}[c]{@{}r@{}}SICK:\\ STS-benchmark:\end{tabular}                                        & \begin{tabular}[c]{@{}c@{}}71.30 / 62.67 / 49.09\\ 68.61 / 62.15 / 49.99\end{tabular}                                                                    & \begin{tabular}[c]{@{}c@{}}68.85\%\\ 72.85\%\end{tabular}            \\ \midrule
			SIF-W2V:                                          & \begin{tabular}[c]{@{}r@{}}SICK:\\ STS-benchmark:\end{tabular}                                        & \begin{tabular}[c]{@{}c@{}}73.44 / 65.93 / 52.60\\ 70.39 / 63.51 / 52.06\end{tabular}                                                                    & \begin{tabular}[c]{@{}c@{}}71.63\%\\ 73.96\%\end{tabular}            \\ \midrule
			USIF-W2V:                                         & \begin{tabular}[c]{@{}r@{}}SICK:\\ STS-benchmark:\end{tabular}                                        & \begin{tabular}[c]{@{}c@{}}73.70 / 66.06 / 52.71\\ 69.95/ 62.85 / 51.11\end{tabular}                                                                     & \begin{tabular}[c]{@{}c@{}}71.51\%\\ 73.07\%\end{tabular}            \\ \midrule
			SUBS-W2V-STOP:                                    & \begin{tabular}[c]{@{}r@{}}SICK:\\ STS-benchmark:\end{tabular}                                        & \begin{tabular}[c]{@{}c@{}}66.10 / 59.28 / 46.94\\ 71.58 / 65.36 / 53.05\end{tabular}                                                                    & \begin{tabular}[c]{@{}c@{}}71.02\%\\ 74.10\%\end{tabular}            \\ \midrule
			INF-FT:                                           & \begin{tabular}[c]{@{}r@{}}SICK:\\ STS-benchmark:\end{tabular}                                        & \begin{tabular}[c]{@{}c@{}}75.94 / 68.95 / 56.56\\ 74.77 / 67.88 / 55.60\end{tabular}                                                                    & \begin{tabular}[c]{@{}c@{}}74.48\%\\ 74.36\%\end{tabular}            \\ \bottomrule
		\end{tabular}%
	}
	\caption{Pearson Correlation Coefficient (PCC) performance ($\times 100$) for SICK and STS-benchmark \textit{dev} and \textit{test} sets when WER is varied ($0\%$, $10\%$, and $30\%$). The last column of each table shows the ratio (as a percentage) of the PCC at $\mathrm{WER} = 30\%$  to the PCC at $\mathrm{WER} = 0\%$ to demonstrate the robustness in STS performance of each sentence embedding to ASR errors at a high WER.}
	\label{table:PCC_WER}
\end{table}

\subsection{Robustness of Sentence Embeddings to Simulated ASR Errors}\label{subsec:embedding_error}
To study the effects of ASR errors on sentence embeddings, we first computed a sentence embedding for each sentence in SICK \cite{marelliSemEval2014TaskEvaluation2014} and STS-benchmark \cite{cerSemEval2017TaskSemantic2017} \emph{dev} and \emph{test} sets using each of the methods described in Section~\ref{sec:embeddings}.
Since \emph{GloVe} embeddings were used to generate the simulated ASR substitution errors, we used \emph{FastText} (for \emph{InferSent}) and \emph{word2vec} embeddings (all other methods) to generate sentence embeddings.
For each method, we corrupted the sentences in the text with a defined WER between $0\%$ and $50\%$ with the simulator described in Section~\ref{sec:asr}.
Then, each sentence in each set is compared with its corrupted counterpart using the relevant similarity metric (\ie cosine or principal angle similarity).

The results are shown in Figure \ref{fig:asr_sentence_embeddings}, in which all methods show a steady linear decline in average similarity between original and corrupted sentences as WER is increased.
As expected, when WER is $0\%$, the sentence embedding similarity is equal to $1$ for all methods.
Simple averaging shows the least significant decline as WER is increased, \ie at $\text{WER} = 50\%$ we see $\text{sim}_{\text{avg}} \approx 0.776$ for unweighted averaging and $\text{sim}_{\text{avg}} \approx 0.742$ for unweighted averaging and stop words removed.
However, we see a significantly steeper decline for SIF and uSIF when $\text{WER} = 50\%$, \ie $\text{sim}_{\text{avg}} \approx 0.592$ for SIF and $\text{sim}_{\text{avg}} \approx 0.633$ for uSIF.
The subspace representation and \emph{InferSent} show a moderate decline in between these two extremes.
These results are in line with our intuition, as we expect word substitution errors to have the smallest overall impact on unweighted average sentence embeddings.
Also as expected, unweighted averages with stop words are more impacted by ASR errors, since stop words in the original corpus could be replaced by content words.
This would lead to a greater difference between original and corrupted sentence similarity scores. 
SIF and uSIF are the most impacted by word substitution errors.
We believe this is explained by the weighted average computation, \ie if a frequent word is replaced by a less frequent word, it may have a greater impact on the overall sentence embedding.
Additionally, it is likely the principal components of the embedding matrix are drastically altered by the introduced error and variance in the dataset, leading to larger differences in sentence embedding representations after corruption and common component removal.
Since the common component removal is weighted by $\lambda_i \leq 1$ for each of the $i$ principal components in uSIF, the overall impact of the introduced variance due to ASR errors is diminished when compared to the single component removal step in SIF.

\begin{figure}[t]
	\centering
	\includegraphics[width=0.85\columnwidth]{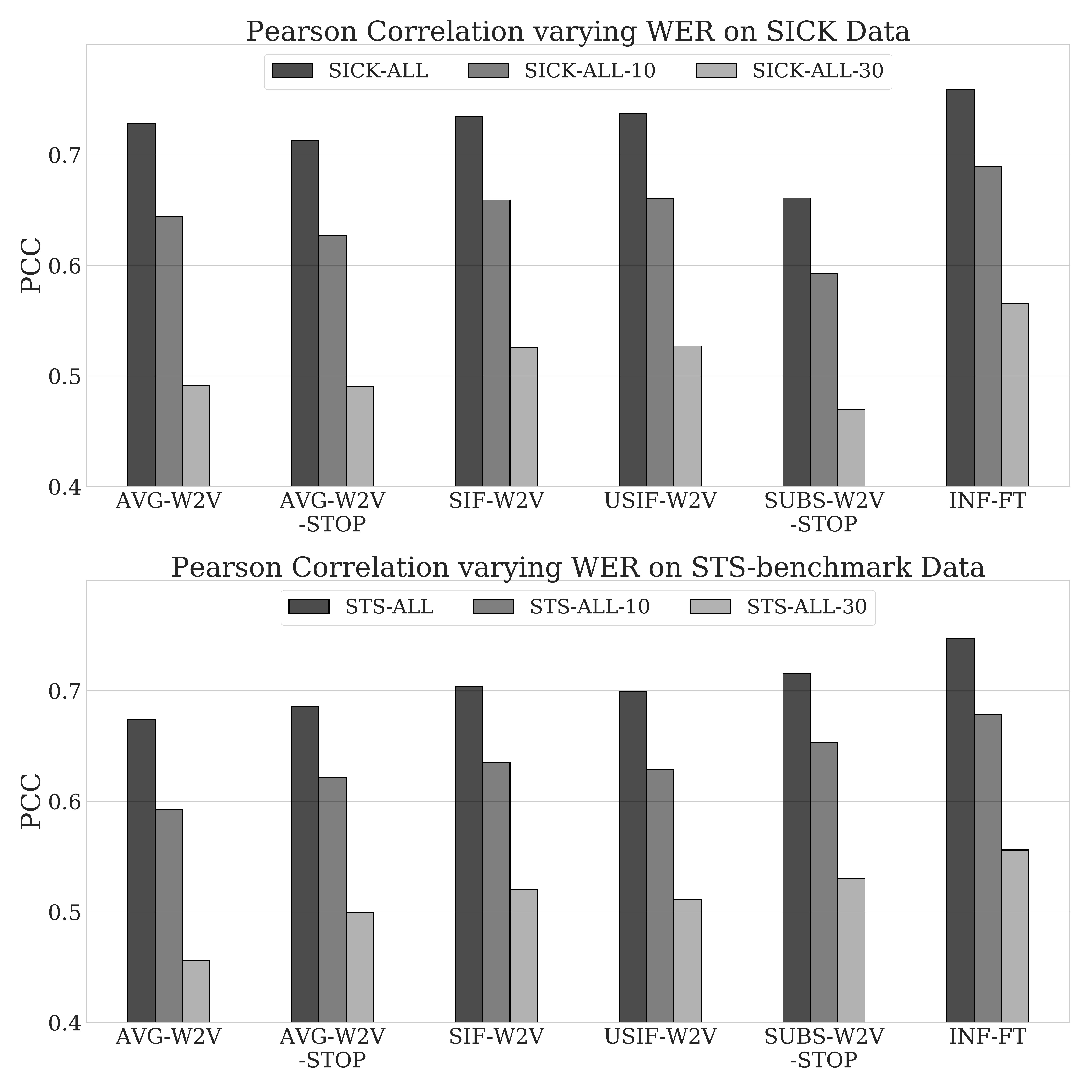}
	\caption{Graphical depiction of the STS performance of various sentence embeddings with simulated word substitution error, see Table~\ref{table:PCC_WER}}
	\label{fig:PCC_WER}
\end{figure}

\subsection{Evaluation of STS Results with Word Substitution Errors}\label{subsec:STS_error}
We next compared the STS performance of the sentence embeddings on the original and corrupted corpora (with $10\%$ and $30\%$ WER) with the \emph{dev} and \emph{test} sets of SICK~\cite{marelliSemEval2014TaskEvaluation2014} and STS-benchmark~\cite{cerSemEval2017TaskSemantic2017}.
The\emph{ Pearson Correlation Coefficient} (PCC) between the computed similarities and the annotated similarity scores in the corpora is the standard metric by which we evaluate STS performance of a given method.
The results are seen in Table~\ref{table:PCC_WER} and Figure~\ref{fig:PCC_WER}.

On the original sentences, simple unweighted averaging provides a strong benchmark for STS tasks on both corpora, with nearly equivalent results when stop words are removed.
In most cases, the weighted average and de-noising provided by SIF and uSIF improve upon the results of unweighted averages, with both methods displaying near-identical performance.
The subspace results are somewhat inconclusive, as they show a slight improvement over averages, SIF, and uSIF on STS-benchmark but a decrease in performance on SICK.
The authors in \cite{muRepresentingSentencesLowRank2017} chose $N=4$ empirically as the subspace rank, based on a variety of corpora which comprise the STS-benchmark set.
It is possible that the absolute performance of the subspace sentence embedding can be improved by tuning the fixed subspace rank for SICK as well.
Unsurprisingly, \emph{InferSent} is consistently the strongest performer, likely due to its supervised training on the SNLI corpus.

When, ASR errors are introduced, the STS performance for each method changes significantly, as evidenced by the results in Table \ref{table:PCC_WER}.
Though the simple averages were least impacted with the introduction of ASR errors (Section~\ref{subsec:embedding_error}), they perform worst among the methods tested on STS tasks with a high WER.
On the other hand, SIF and uSIF embeddings were most impacted by ASR errors but perform among the best in STS when the WER is high. 
Again, we suspect this is due to the common component removal steps in SIF and uSIF, which effectively act as de-noising steps removing some of the additional variance in the embedding matrix due to substitution errors.
Since SIF and uSIF display near-identical STS performance across both corpora, we think uSIF may be a slightly better choice due to its increased robustness to ASR errors.
Also, as suspected, we see that the subspace embeddings show increased STS performance robustness to word substitution errors when compared to averages if we consider the PCC ratio between high WER (30\%) and original sentences.
Subspace embeddings slightly outperform SIF and uSIF on STS-benchmark and slightly under-perform SIF and uSIF on SICK by the same metric.
Again, \emph{InferSent} not only shows the best absolute performance on the original sentences, but shows the best performance with a high WER rate as well.

\section{Conclusion}\label{sec:conclusion}
In this paper, we introduced a simulator that automates word substitution errors (given a WER) on perfectly transcribed corpora to simulate ASR-plausible errors, considering both phonemic and semantic similarities between words.
We then used the simulator to intentionally corrupt standard corpora used for textual similarity tasks (SICK~\cite{marelliSICKCureEvaluation} and STS-benchmark~\cite{cerSemEval2017TaskSemantic2017}).
From this, we were able to evaluate the impact that word substitution errors may have on some of the most recently developed techniques for sentence embeddings.
We also evaluated the STS performance of each of these sentence embedding methods after introducing substitution errors with our simulator.
We found several interesting results.
For example, average sentence embeddings perform well for perfectly transcribed text, but show poorer STS performance when errors are introduced if compared to more advanced methods.
On the other hand, pre-trained encoders, such as \emph{InferSent} not only show state-of-the-art performance on STS tasks with perfectly transcribed text, but also seem to show increased robustness to error for STS performance.
If it is not possible to use an encoder like \emph{InferSent}, the weighted average and smoothing provided by SIF/uSIF or the low-rank subspace representation by Mu \etal \cite{muRepresentingSentencesLowRank2017} seem to be reasonable improvements over simple averages when it comes to STS performance for high-WER transcriptions.

\bibliographystyle{references/IEEEbib}
\bibliography{references/asr_error_refs,references/NLP_refs}

\end{document}